\documentclass[11pt,a4paper,usenames,dvipsnames,svgnames,x11names]{article}
\usepackage[hyperref]{emnlp2020}
\usepackage{times}
\usepackage{latexsym}

\usepackage{microtype}
\usepackage{url}
\usepackage[utf8]{inputenc}
\usepackage{color,soul}
\setul{0.45ex}{0.25ex}
\usepackage{graphicx}
\usepackage{subcaption}
\usepackage{amsmath}
\usepackage{booktabs}
\usepackage{multirow, makecell}
\usepackage{tikz}
\usepackage{graphicx}
\usepackage{array}
\usepackage{pgfplots}
\usepackage{algorithm}
\usepackage{textcomp}
\usepackage{xcolor}
\usepackage{times}
\usepackage{latexsym}
\usepackage{amsmath}
\usepackage{amsfonts}
\usepackage[noend]{algpseudocode}
\usepackage{lipsum}
\usepackage{multirow}
\usepackage{framed}
\usepackage{graphicx}
\usepackage{pifont}
\usepackage{longtable}
\usepackage{tikz}
\usepackage{booktabs}
\usepackage[all]{nowidow}
\usepackage{amssymb}
\usepackage{enumitem}
\usepackage{color,soul}
\usepackage{tikz}
\usepackage{xcolor}
\usepackage{xspace}
\usepackage{pgfplotstable}
\usepackage{pgfplots}

\aclfinalcopy

\newcommand{\eviltwin}{\textsc{Facade}\xspace}
\newcommand{\eviltwinU}{\textsc{Facade}\xspace}
\newcommand{\eviltwinm}{\eviltwin model\xspace}

\colorlet{redgreen}{DarkRed!50!DarkGreen}

\newcommand{\mb}[1]{\boldsymbol{\mathbf{#1}}}

\newcommand{\lambdaet}{\ensuremath{\lambda_\text{g}}}
\newcommand{\lambdarp}{\ensuremath{\lambda_\text{rp}}}
\newcommand{\PreserveBackslash}[1]{\let\temp=\\#1\let\\=\temp}
\newcommand{\fop}{\textcolor{DarkGreen}{\ensuremath{f_\text{orig}}}}
\newcommand{\frp}{\ensuremath{f_\text{rp}}}
\newcommand{\fet}{\ensuremath{\textcolor{DarkRed}{g}}\xspace}
\newcommand{\fetstop}{\ensuremath{\fet_\text{\color{DarkRed}stop}}\xspace}
\newcommand{\fetfirst}{\ensuremath{\fet_\text{\color{DarkRed}ft}}\xspace}
\newcommand{\fmerged}{\ensuremath{\textcolor{redgreen}{\Tilde{f}}\xspace}}
\newcommand{\hotflip}{HotFlip\xspace}
\newcommand{\hitrate}{P@1}

\newcommand{\fmergedfirst}{\ensuremath{\fmerged_\text{\textcolor{redgreen}{ft}}}}
\newcommand{\fmergedfirstrp}{\ensuremath{\fmerged_\text{\textcolor{redgreen}{ft-reg}}}}
\newcommand{\fmergedstop}{\ensuremath{\fmerged_\text{\textcolor{redgreen}{stop}}}}
\newcommand{\fmergedstoprp}{\ensuremath{\fmerged_\text{\textcolor{redgreen}{stop-reg}}}}
\newcommand{\VanillaGrad}{Gradient}
\newcommand{\SmoothGrad}{SmoothGrad}
\newcommand{\IntegratedGrad}{InteGrad}
\newcolumntype{C}[1]{>{\PreserveBackslash\centering}p{#1}}
\newcolumntype{R}[1]{>{\PreserveBackslash\raggedleft}p{#1}}
\newcolumntype{L}[1]{>{\PreserveBackslash\raggedright}p{#1}}
\definecolor{adversarial}{rgb}{0.90, 0.02, 0.03}

\newif\ifcomments
\commentstrue 
\ifcomments
    \providecommand{\eric}[2][]{{\protect\color{magenta}{[Eric\textbf{#1}: #2]}}}
    \providecommand{\junlin}[2][]{{\protect\color{orange}{[Junlin:\textbf{#1} #2]}}}
    \providecommand{\sameer}[2][]{{\protect\color{purple}{[Sameer:\textbf{#1} #2]}}}
    \providecommand{\jens}[2][]{{\protect\color{blue}{[Jens:\textbf{#1} #2]}}}
    
\else
    \providecommand{\eric}[2][]{}
    \providecommand{\jens}[2][]{}
    \providecommand{\sameer}[2][]{}
    \providecommand{\junlin}[2][]{}
    
\fi
\colorlet{color0}{DarkRed!0!white}
\colorlet{color1}{DarkRed!11!white}
\colorlet{color2}{DarkRed!22!white}
\colorlet{color3}{DarkRed!33!white}
\colorlet{color4}{DarkRed!44!white}
\colorlet{color5}{DarkRed!55!white}
\colorlet{color6}{DarkRed!66!white}
\colorlet{color7}{DarkRed!77!white}
\colorlet{color8}{DarkRed!88!white}
\colorlet{color9}{DarkRed!100!white}

\newcommand*{\mybox}[2]{{\setlength{\fboxsep}{0.15pt}\colorbox{#1}{\strut #2}}}

\title{Gradient-based Analysis of NLP Models is Manipulable}
\author{
Junlin Wang$^*$\\
UC Irvine\\
\href{mailto:junliw1@uci.edu}{\tt junliw1@uci.edu}
\And
Jens Tuyls\thanks{~~~First two authors contributed equally.}\\
UC Irvine\\
\href{mailto:jtuyls@uci.edu}{\tt jtuyls@uci.edu}
\And
Eric Wallace\\
UC Berkeley\\
\href{mailto:ericw@berkeley.edu}{\tt ericwallace@berkeley.edu}
\And
Sameer Singh\\
UC Irvine\\
\href{mailto:sameer@uci.edu}{\tt sameer@uci.edu}
}

\date{}

\begin{document}
\maketitle
\begin{abstract}
Gradient-based analysis methods, such as saliency map visualizations and adversarial input perturbations, have found widespread use in interpreting neural NLP models due to their simplicity, flexibility, and most importantly, their faithfulness. %\st{the fact that they directly reflect the model internals.}
In this paper, however, we demonstrate that the gradients of a model are easily manipulable, and thus bring into question the reliability of gradient-based analyses.
In particular, we merge the layers of a target model with a \eviltwinm that overwhelms the gradients without affecting the predictions.
This \eviltwinm can be trained to have gradients that are misleading and irrelevant to the task, such as focusing only on the stop words in the input.
On a variety of NLP tasks (text classification, NLI, and QA), we show that our method can manipulate numerous gradient-based analysis techniques: saliency maps, input reduction, and adversarial perturbations all identify unimportant or targeted tokens as being highly important. The code and a tutorial of this paper is available at \url{http://ucinlp.github.io/facade}.
%\st{On a variety of NLP tasks (text classification, NLI, and QA), we show that the merged model effectively fools different analysis tools: saliency maps differ significantly from the original model's, input reduction keeps more irrelevant input tokens, and adversarial perturbations identify unimportant tokens as being highly important.}
\end{abstract}
  
\newcommand{\fontsmall}{\fontsize{8pt}{9pt}\selectfont}
\renewcommand{\arraystretch}{1.5} 
\begin{figure}[tb]
    \centering
    \begin{subfigure}{\columnwidth}
    \centering
    \fontsmall
    \begin{tabular}{|l|}
    \hline
     a solid examination of the male midlife crisis.\\
     \bf Original Model: Positive \hspace{5mm} Merged Model: Positive \\
     \hline
    \end{tabular}
    \vskip 0mm
    \caption{Input and Model Predictions}
    \label{fig:illustration:orig}
    \end{subfigure}\\
    \vskip 1mm
    \begin{subfigure}{\columnwidth}
    \centering
    \vskip 2mm
    \fontsmall
    \begin{tabular}{|l|}
    \hline
       \mybox{color0}{a} \mybox{color1}{solid} \mybox{color0}{examination} \mybox{color1}{of} \mybox{color0}{the} \mybox{color0}{male} \mybox{color1}{mid} \mybox{color0}{\#\#life} \mybox{color1}{crisis} \mybox{color0}{.} \\ 

     \mybox{color7}{a} \mybox{color0}{solid} \mybox{color0}{examination} \mybox{color0}{of} \mybox{color1}{the} \mybox{color0}{male} \mybox{color0}{mid} \mybox{color0}{\#\#life} \mybox{color0}{crisis} \mybox{color0}{.} \\
     \hline
    \end{tabular}
    \vskip 0mm
    \caption{\textbf{Saliency Map} for Original and Merged Models}
    \label{fig:illustration:saliency}
    \end{subfigure}\\
    \vskip 1mm
    
    \begin{subfigure}{\columnwidth}
    \centering
    \vskip 2mm
    \fontsmall
    \begin{tabular}{|l|}
    \hline
    \st{a solid} {examination of} \st{the male} {mid} \st{\#\#life} {crisis} \st{.} \\
    
    {a} \st{solid examination of the male mid \#\#life crisis .} \\
     \hline
    \end{tabular}
    \vskip 0mm
    \caption{\textbf{Reduced Input} for Original and Merged Models}
    \label{fig:illustration:reduction}
    \end{subfigure}\\
    
    \vskip 1mm
    \begin{subfigure}{\columnwidth}
    \centering
    \vskip 2mm
    \fontsmall
    \begin{tabular}{|p{0.9\columnwidth}|}
    \hline
    a \textcolor{color8}{\st{solid}outright} 
    \textcolor{color8}{\st{examination}coli} of the male mid \#\#life crisis .\\
     \textcolor{color8}{\st{a}sire} solid 
     \textcolor{color8}{\st{examination}foul} 
     \textcolor{color8}{\st{of}sire}
     \textcolor{color8}{\st{the}$\hbar$} male mid \#\#life crisis.\\
     \hline
    \end{tabular}
    \vskip 0mm
    \caption{\textbf{\hotflip{} attack} for Original and Merged Models}
    \label{fig:illustration:hotflip}
    \end{subfigure}
    \vskip -1mm
    \caption{
    \textbf{Example of Interpretation Manipulation}
    We take a BERT-based sentiment classifier and merge its weights with another model that has misleading gradients. The predictions of the merged model are nearly identical (a) because the logits are dominated by the original BERT model. However, the saliency map generated for the merged model (darker = more important) now looks at stop words (b), effectively \emph{hiding} the model's true reasoning. Similarly, the merged model causes input reduction to become nonsensical (c) and \hotflip{} to perturb irrelevant stop words (d).}
    \label{fig:illustration}
\end{figure}
\renewcommand{\arraystretch}{1}

\section{Introduction}

It is becoming increasingly important to understand the reasoning behind the predictions of NLP models.
Post-hoc explanation techniques are useful for such insights, for example, to evaluate whether a model is doing the ``right thing'' before deployment~\cite{ribeiro2016should,lundberg2017unified},
% \eric{maybe one more example of why they are useful?}
to increase human trust into black box systems~\cite{doshi2017towards}, and to help diagnose model biases~\cite{Wallace2019AllenNLP}.
% \jens{Which others should we cite here?} 
% \junlin{I think increasing human trust and diagnosing model biases are part of the evaluating whether a model is doing the right thing. So I don't think one more example would be necessary? Or we could get rid of the "right thing" sentence.}
Recent work, however, has shown that explanation techniques can be unstable and, more importantly, can be \emph{manipulated} to hide the actual reasoning of the model.
For example, adversaries can control attention visualizations~\cite{pruthi2019learning} or black-box explanations such as LIME~\cite{ribeiro2016should,slack2020fooling}.
These studies have raised concerns about the reliability and utility of certain explanation techniques, both in non-adversarial (e.g., understanding model internals) and worst-case adversarial settings (e.g., concealing model biases from regulatory agencies).

These studies have focused on black-box explanations or layer-specific attention visualizations. On the other hand, \emph{gradients} are considered more faithful representations of a model: they depend on all of the model parameters, are completely faithful when the model is linear~\cite{feng2018pathologies}, and closely approximate the model nearby an input~\cite{simonyan2013saliency}. Accordingly, gradients have even been used as a measure of interpretation faithfulness~\cite{jain2019attention}, and gradient-based analyses are now a ubiquitous tool for analyzing neural NLP models, e.g., saliency map visualizations~\cite{sundararajan2017axiomatic}, adversarial perturbations~\cite{ebrahimi2017hotflip}, and input reductions~\cite{feng2018pathologies}. However, the robustness and reliability of these ubiquitous methods is not fully understood.

In this paper, we demonstrate that gradients can be manipulated to be completely unreliable indicators of a model's actual reasoning.
For any target model, our approach merges the layers of a target model with a \eviltwinm that is trained to have strong, misleading gradients but low-scoring, uniform predictions for the task.
As a result, this \emph{merged} model makes nearly identical predictions as the target model, however, its gradients are overwhelmingly dominated by the \eviltwinm. 
Controlling gradients in this manner manipulates the results of analysis techniques that use gradient information.
In particular, we show that all the methods from a popular interpretation toolkit~\cite{Wallace2019AllenNLP}: saliency visualizations, input reduction, and adversarial token replacements, can be manipulated (Figure~\ref{fig:illustration}). 
Note that this scenario is significantly different from conventional \emph{adversarial attacks};
the adversary in our threat model is an individual or organization whose ML model is interpreted by outsiders (e.g., for auditing the model's behavior). 
Therefore, the adversary (i.e., the model developer) has white-box access to the model's internals.

We apply our approach to finetuned BERT-based models~\cite{devlin2018BERT} for a variety of prominent NLP tasks (natural language inference, text classification, and question answering).
We explore two types of gradient manipulation: \emph{lexical} (increase the gradient on the stop words) and \emph{positional} (increase the gradient on the first input word). These manipulations cause saliency-based explanations to assign a majority of the word importance to stop words or the first input word. Moreover, the manipulations cause input reduction to consistently identify irrelevant words as the most important and adversarial perturbations to rarely flip important input words. Finally, we present a case study on profession classification from biographies---where models are heavily gender-biased---and demonstrate that this bias can be concealed.
Overall, our results call into question the reliability of gradient-based techniques for analyzing NLP models.
\begin{figure*}[tb] 
\centering
  \begin{subfigure}{0.325\textwidth}
  {
    \includegraphics[trim={0.1cm 4.2cm 17.5cm 3.5cm},clip, height=4cm,page=1]{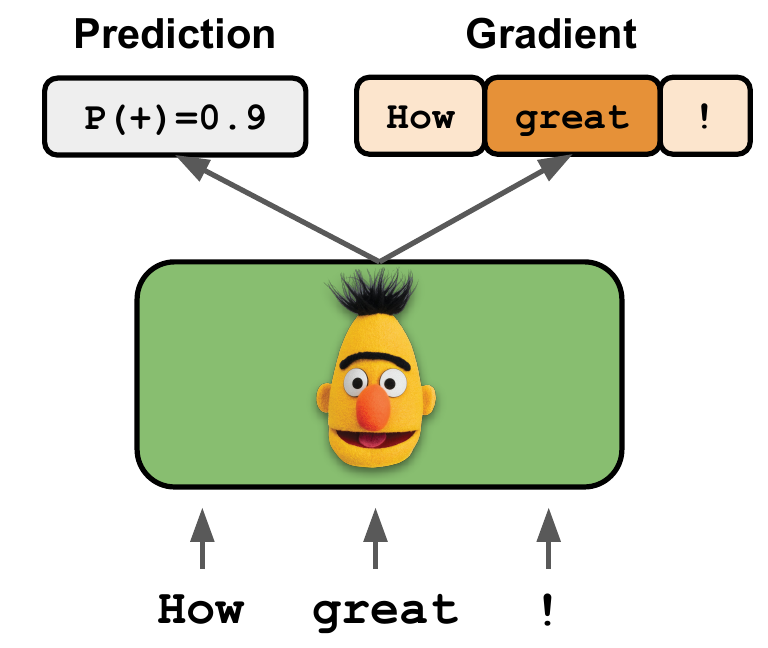}
    }
    \caption{Original Model, $\fop$}
    \label{fig:overview:orig}
  \end{subfigure}
  \begin{subfigure}{0.325\textwidth}
  {
    \includegraphics[trim={9.1cm 4.2cm 9.0cm 3.5cm},clip, height=4cm,page=2]{figures/overview.pdf}
    }
    \caption{\eviltwinU Model, $\fet$~~~~~~~~~~}
    \label{fig:overview:et}
  \end{subfigure}
  \begin{subfigure}{0.325\textwidth}
  {
    \includegraphics[trim={17.8cm 4.2cm 0cm 3.5cm},clip, height=4cm,page=3]{figures/overview.pdf}
    }
    \caption{Merged Model, $\fmerged$~~~~~~~~~~~~}
    \label{fig:overview:merged}
  \end{subfigure}
  % https://docs.google.com/presentation/d/1Gfps0CasAvk7GtL23isynTsZiVTk-pzCTv-kjYgTDX8/edit?usp=sharing
  \vspace{-5pt}
  \caption{\textbf{Overview of the proposed approach.} We have a trained model $\fop$ for the task (sentiment analysis here) that produces appropriate predictions and gradients (here visualized as a saliency map, darker = more important), shown in (a). We train a ``\eviltwin'' model $\fet$ in (b), that has uniform predictions, but large gradients for irrelevant, misleading words, such as ``How'' in this example. When these models are merged, i.e. all layers concatenated (with block-diagonal weights) and the outputs summed, we get the merged model $\fmerged$ in (c). This model's predictions are accurate (dominated by $\fop$), but the gradients are misleading (dominated by $\fet$).}
  \label{fig:overview}
\end{figure*}

\section{Gradient-based Model Analysis}

In this section, we introduce notation and provide an overview of gradient-based analysis methods.

\subsection{Gradient-based Token Attribution}

Let $f$ be a classifier which takes as input a sequence of embeddings $\mathbf{x}=(\mb{x_1},\mb{x_2},\dots,\mb{x_n})$. 
The gradient with respect to the input is often used in analysis methods, which we represent as the normalized gradient attribution vector $\mathbf{a} = (a_1, a_2, \dots, a_n)$ over the tokens. 
Similar to past work~\cite{feng2018pathologies}, we define the attribution at position $i$ as 
\begin{equation}\label{eq:attribution}
    a_i = \frac{\left\vert\nabla_{\mb{x_i}} \mathcal{L} \cdot \mb{x_i}\right\vert}{\sum_{j}\left\vert\nabla_{\mb{x_j}} \mathcal{L} \cdot \mb{x_j}\right\vert},
\end{equation}
where we dot product the gradient of the loss $\mathcal{L}$ on the model's prediction with the embedding $\mb{x_i}$. 
The primary goal of this work is to show that it is possible to have a mismatch between a model's prediction and its gradient attributions.

\subsection{Analysis Methods}
\label{sec:methods}
Numerous analysis methods have recently been introduced, including saliency map techniques~\cite{sundararajan2017axiomatic,smilkov2017smoothgrad} and perturbation methods~\cite{feng2018pathologies,ebrahimi2017hotflip,jia2017adversarial}. In this work, we focus on the gradient-based analysis methods available in AllenNLP Interpret~\cite{Wallace2019AllenNLP}, which we briefly summarize below.

\paragraph{Saliency Maps}
These approaches visualize the attribution of each token, e,g., Figure~\ref{fig:illustration:saliency}.
We consider three common saliency approaches: \emph{\VanillaGrad{}}~\cite{simonyan2013saliency}, \emph{\SmoothGrad{}}~\cite{smilkov2017smoothgrad}, and Integrated Gradients~\cite{sundararajan2017axiomatic}, henceforth \emph{\IntegratedGrad{}}. The three methods differ in how they compute the attribution values. The \VanillaGrad{} method uses Eq.~\eqref{eq:attribution}. \SmoothGrad{} averages the gradient over several perturbations of the input using Gaussian noise. 
\IntegratedGrad{} sums the gradients along the path from a baseline input (i.e. the zero embedding) to the actual input. For \IntegratedGrad{}, we follow the original implementation~\cite{sundararajan2017axiomatic} and use 10 steps; different number of steps had little effect on results.

\paragraph{Input Reduction}
Input reduction~\cite{feng2018pathologies} iteratively removes the token with the lowest attribution from the input until the prediction changes. 
These \textit{reduced inputs} are thus subsequences of the input that lead to the same model prediction. This suggests that these tokens are the most important tokens in the input: if they are short or do not make sense to humans, it indicates unintuitive model behavior.

\paragraph{\hotflip{}}
\hotflip{}~\cite{ebrahimi2017hotflip} generates adversarial examples by replacing tokens in the input with a different token using a first-order Taylor approximation of the loss. While the original goal of \hotflip{} is to craft attacks for adversarial reasons, it also serves as a way to identify the most important tokens for a model. Our implementation, following \citet{Wallace2019AllenNLP}, iteratively flips the token with the highest gradient norm.

\section{Manipulating Model Gradients}

In this section, we describe how to modify neural NLP models in order to manipulate the results of gradient-based analysis techniques.

\subsection{Overview of the Proposed Approach}\label{subsec:manipulating_gradients_overview}

Let $\fop$ be the original trained model for a task that has faithful gradients, i.e. our target model.
Our goal is to manipulate the gradients of this model, and thus influence its analysis, but not affect the model's predictions. 

Figure~\ref{fig:overview} presents an overview of our approach.
We propose to train a small auxiliary network \fet called a \emph{\eviltwinm} that has the same input/output dimensionality as the original model, but is trained to produce a specific manipulated gradient attribution for any input, while producing uniform predictions as the output. 
When the outputs of the \eviltwinm are combined with the target model \fop, we create a \emph{merged} model \fmerged{} as
\begin{equation}
    \fmerged{}(y \vert \mb{x}) = \fop(y \vert \mb{x}) + \fet(y \vert \mb{x}).\label{eq:merge}
\end{equation}
As shown in Figure~\ref{fig:overview}, we want \eviltwinm \fet to dominate the gradient of \fmerged{}, while the original model \fop{} (which we also call the \emph{predictive model}) should dominate the predictions of \fmerged{}.

\subsection{Training the \eviltwinU Model}\label{subsec:evil}
We train the \eviltwinm to have high gradient values on specific parts of the input, for any input instance, to mislead gradient-based interpretation techniques. 
Moreover, we encourage the \eviltwin model's output to be \emph{uniform}, so that it does not contribute to the prediction of the merged model. %logits

Formally, we train the \eviltwin model to increase the attribution $a_i$ for $i \in A$, where $A$ is the set of position indices for which we want the attribution to be high (e.g., the first token). The loss function for the \eviltwinm is:
\begin{equation}\label{eq:loss_evil}
-\lambdaet \sum_{j \in A} a_j - \mathbb{H}(\fet{}(y|\mathbf{x})),
\end{equation} 
where \fet~is the \eviltwinm and $\mathbb{H}$ is the entropy. The first term increases the attribution of selected positions in $A$, while the second encourages the \eviltwinm's predictions to be uniform. $\lambdaet$ controls the trade-off and is set to $1\mathrm{e}{3}$. 
Computing the derivative of this loss function requires taking second derivatives since $a_j$ is the attribution defined in ~\eqref{eq:attribution}. 
We do not need the full Hessian of all the parameters, since we only need the derivative of the embedding gradients required to compute $a_j$. Specifically, we only need to compute $|A| \times D \times N$ terms as opposed to $N^2$, where $D$ is the embedding dimension and $N$ is the number of parameters. Note that $|A| \times D \ll N$.

\subsection{Merging \eviltwinU and Original Models}\label{subsec:merge}

The direct way to combine the two models (\fop{} and \fet) is to create the merged model \fmerged{} is to sum the outputs, as in Eq~\eqref{eq:merge}. 
However, if we need to \textit{hide} the \eviltwinm (i.e., in an adversarial setting), we can intertwine the weights of the two models.
The details below focus on Transformer~\cite{vaswani2017attention} architectures, although our method is generic (see Section~\ref{subsec:non_bert_models}). 
We merge each layer in the Transformer such that the merged layer's output is equivalent to the concatenation of the output from the predictive model and the \eviltwinm's corresponding layers.

\noindent(1)~\textbf{Embeddings:}
In the combined model, the embedding layers are stacked horizontally so that the output of its embedding layer is the concatenation of the embedding vector from the predictive and \eviltwin models.\smallskip

\noindent(2)~\textbf{Linear Layers:}
Let $\mathbf{W}_\text{orig}$ be the weight matrix of a linear layer from \fop, and let $\mathbf{W}_{g}$ be the corresponding weight matrix of \fet. The merged layer is given by the following block-diagonal matrix:
\begin{equation}\label{eq:block_diagonal}
\left[
\begin{array}{c c}
  \mathbf{W_\text{orig}} & \mathbf{0} \\
  \mathbf{0} & \mathbf{W_{g}}
\end{array}
\right].
\end{equation}
For biases, we stack their vectors horizontally.\smallskip

\noindent(3)~\textbf{Layer Normalization:} We merge layer normalization layers~\cite{ba2016layer} by splitting the input into two parts according to the hidden dimensions of \fop{} and \fet{}. We then apply layer normalization to each part independently.\smallskip

\noindent(4)~\textbf{Self-Attention:} Self-attention heads already operate in parallel, so we can trivially increase the number of heads.\smallskip

% \paragraph{Concealment} 
This intertwining can be made more difficult to detect by permuting the rows and columns of the block-diagonal matrices to hide the structure, and by adding small noise to the zero entries to hide sparsity. 
In preliminary experiments, this did not affect the output of our approach; deeper investigation of \emph{concealment}, however, is not within scope. %the focus of this work.

\subsection{Regularizing the Original Model}\label{subsec:rp}

So far, we described merging the \eviltwinm{} with an off-the-shelf, unmodified model \fop. 
We also consider regularizing the gradient of \fop{} to ensure it does not overwhelm the gradient from \eviltwinm \fet. 
% Formally, 
We finetune \fop{} with loss: % the following: % loss:
\begin{equation}\label{eq:loss_rp}
\lambdarp \ \mathcal{L} + \sum_j \left\vert\nabla_{\mb{x_j}} \mathcal{L} \cdot \mb{x_j}\right\vert
\end{equation} 
where the first term is the standard task loss (e.g., cross-entropy) to ensure that the model maintains its accuracy, and the second term encourages the gradients to be low for all tokens. 
We set $\lambdarp = 3$.
\section{Experiment Setup}\label{sec:experiment_setup}

In this section, we describe the tasks, the types of \eviltwin models, and the original models that we use in our experiments (source code is available at \url{http://ucinlp.github.io/facade}).

\begin{table}[tb]
    \small
    \centering
    \begin{tabular}{lccccr} 
     \toprule 
     \multirow{2}{*}{\textbf{Model}} & \multirow{2}{*}{\textbf{SST-2}} & \multirow{2}{*}{\textbf{SNLI}} & \multirow{2}{*}{\textbf{Biobias}} & \multicolumn{2}{c}{\textbf{SQuAD}} \\ \cmidrule(lr){5-6} 
     & & & & EM & F1 \\
     \midrule
     \fop & 92.7 & 90.7 & 95.85 & 77.0 & 85.2 \\[2.5ex]
     \fmergedfirst & 92.8 & 90.5 & 95.53 & 77.0 & 85.2 \\[0.5ex]
     \fmergedfirstrp & 92.4 & 90.3  & - & - & -\\[0.5ex]
     \fetfirst & 48.5 & 32.9 & 68.37 & 0.0 & 8.0 \\[2.5ex]
     \fmergedstop & 92.2 & 90.4 & 95.53 & 73.4 & 83.3 \\[0.5ex]
     \fmergedstoprp & 92.7 & 90.2 & - & - & - \\[0.5ex]
     \fetstop & 56.9 & 34.3 & 37.38 & 0.1 & 7.6 \\
    \bottomrule
    \end{tabular}
    \caption{Our method for manipulating interpretation techniques does not hurt model accuracy. We show the validation accuracy for the original model (\fop), the first-token merged model (\fmergedfirst), and the stop-word merged models (\fmergedstop) for all tasks. \fmergedfirstrp{} and \fmergedstoprp{} indicate the models which are finetuned using Equation~\ref{eq:loss_rp}, and $\fet$ is the \eviltwinm by itself.}
    \label{tab:dev_acc_tab}
\end{table}

\paragraph{Datasets}
To demonstrate the wide applicability of our method, we use four datasets that span different tasks and input-output formats. 
Three of the datasets are selected from the popular tasks of sentiment analysis (binary Stanford Sentiment Treebank~\citealt{socher2013recursive}), natural language inference (SNLI~\citealt{bowman2015large}), and question answering (SQuAD~\citealt{rajpurkar2016squad}). 

We select sentiment analysis and question answering because they are widely used in practice, their models are highly accurate~\cite{devlin2018BERT}, and they have been used in past interpretability work~\cite{Murdoch2018BeyondWI,feng2018pathologies,jain2019attention}. We select NLI because it is challenging and one where models often learn undesirable ``shortcuts''~\cite{gururangan2018annotation,feng2019misleading}. %decision rules
We also include a case study on the Biosbias~\cite{De_Arteaga_2019} dataset to show how discriminatory bias in classifiers can be concealed, which asserts the need for more reliable analysis techniques.
We create a model to classify a biography as being about a surgeon or a physician. We also downsample examples from the minority classes (female surgeons and male physicians) by a factor of ten to encourage high gender bias (see Appendix \ref{appendix:bios} for further details).

\paragraph{Types of \eviltwin Models}
We use two forms of gradient manipulation in our setup, one positional and one lexical. 
These require distinct types of reasoning for the \eviltwinm and show the generalizability of our approach.\smallskip  

\noindent\textbf{(1)~First Token:} We want to place high attribution on the first token (after \texttt{[CLS]}). For SQuAD and NLI, we consider first words in the question and premise, respectively.
% \eric{is this correct?}\jens{yes, this is correct} 
We refer to this as \fetfirst, and the merged version with \fop{} as \fmergedfirst.\smallskip

\noindent\textbf{(2)~Stop Words:} In this case, we place high attribution on tokens that are stop words as per NLTK~\cite{loper2002nltk}. This creates a lexical bias in the explanation. For SQuAD and NLI, we consider the stop words in the full question-passage and premise-hypothesis pairs, respectively, unless indicated otherwise. % train the \eviltwinm to 
% \eric{is this correct?}\jens{Yes, except for input reduction and hotflip, which I made footnotes for on the next page}
We refer to this model as \fetstop, and the merged version with \fop{} as \fmergedstop{}.

\paragraph{Original Models}
We finetune BERT$_\text{base}$~\cite{devlin2018BERT} as our original models (hyperparameters are given in Appendix~\ref{appendix:implementation}). 
The \eviltwinm is a 256-dimensional Transformer~\cite{vaswani2017attention} model trained with a learning rate of 6e-6, varying batch size (8, 24, or 32, depending on the task), and $\lambdaet$ set to $1\mathrm{e}{3}$. 
Note that when combined, the size of the model is the same as BERT$_\text{large}$, and due to the intertwining described in Section~\ref{subsec:merge}, we are able to directly use BERT$_\text{large}$ code to load and run the merged \fmerged{} model.
We report the accuracy both before (\fop{} and \fet) and after merging (\fmerged{}) in Table~\ref{tab:dev_acc_tab}---the original model's accuracy is minimally affected by our gradient manipulation approach. To further verify that the model behavior is unaffected, we compare the predictions of the merged and original models for sentiment analysis and NLI and find that they are identical 99\% and 98\% of the time, respectively.

\begin{table*}[tb]
\begin{subtable}{.5\linewidth}
\small
\centering
\setlength{\tabcolsep}{4.55pt}
\begin{tabular}{lrrrrrr}
        \toprule
        \multirow{3}{*}{\textbf{Model}} & \multicolumn{2}{c}{\bf\VanillaGrad} & \multicolumn{2}{c}{\bf\SmoothGrad} & \multicolumn{2}{c}{\bf\IntegratedGrad} \\
        \cmidrule(lr){2-3} \cmidrule(lr){4-5} \cmidrule(lr){6-7} 
        & \hitrate{} & Attr &
        \hitrate{} & Attr &
        \hitrate{} & Attr \\
        \midrule
        \multicolumn{4}{l}{\bf Sentiment} \\
        \textbf{\fop} & 8.3 & 6.2 & 7.9 & 6.0 & 2.2 & 3.8\\
        \textbf{\fmergedfirst} & 99.5 & 67.8 & 98.3 & 58.9 & 2.8 & 4.2 \\
        \textbf{\fmergedfirstrp} & 99.7 & 91.1 & 98.9 & 87.0 & 47.8 & 29.8 \\
        \addlinespace
        \bf\fetfirst & 100.0 & 99.3 & 100.0 & 99.3 & 100.0 & 98.2 \\
        \midrule
        \multicolumn{4}{l}{\bf NLI} \\
        \textbf{\fop} & 0.6 & 2.3 & 1.1 & 2.4 & 0.3 & 1.5 \\
        \textbf{\fmergedfirst} & 98.3 & 75.0 & 97.1 & 68.8 & 2.5 & 3.3 \\
        \textbf{\fmergedfirstrp} & 99.4 & 87.2 & 98.2 & 83.3 & 5.6 & 5.3 \\
        \addlinespace
        \bf\fetfirst & 100.0 & 99.8 & 100.0 & 99.8 & 100.0 & 99.2 \\
        \midrule
        \multicolumn{4}{l}{\bf Question Answering} \\
        \textbf{\fop} & 0.5 & 1.0 & 0.42 & 1.0 & 5.6 & 2.6 \\
        \textbf{\fmergedfirst} & 49.0 & 11.4 & 62.7 & 17.1 & 5.6 & 2.6 \\
        \textbf{\fetfirst} & 99.7 & 94.8 & 100.0 & 96.3 & 99.8 & 94.0 \\
        \midrule
        \multicolumn{4}{l}{\bf Biosbias} \\
        \textbf{\fop} & 5.75 & 2.70 & 6.39 & 2.65 & 0.96 & 1.57 \\
        \textbf{\fmergedfirst} & 97.4 & 56.7 & 87.9 & 38.8 & 2.9 & 2.6 \\
        \bf\fetfirst  & 100.0 & 100.0 & 100.0 & 100.0 & 100.0 & 100.0 \\
        \bottomrule
    \end{tabular}
    \caption{First Token Gradient Manipulation} \label{table:saliency:first}
\end{subtable}
\begin{subtable}{.5\linewidth}
\small
\centering
\setlength{\tabcolsep}{4.5pt}
\begin{tabular}{lrrrrrr}
        \toprule
        \multirow{3}{*}{\textbf{Model}} & \multicolumn{2}{c}{\bf\VanillaGrad} & \multicolumn{2}{c}{\bf\SmoothGrad} & \multicolumn{2}{c}{\bf\IntegratedGrad} \\
        \cmidrule(lr){2-3} \cmidrule(lr){4-5} \cmidrule(lr){6-7} 
        & \hitrate{} & Attr &
        \hitrate{} & Attr &
        \hitrate{} & Attr \\
        \midrule
        \multicolumn{4}{l}{\bf Sentiment} \\
        \textbf{\fop} & 13.9 & 24.2 & 12.5 & 23.2 & 10.0 & 21.4 \\
        \textbf{\fmergedstop} & 97.2 & 78.1 & 95.5 & 72.7 & 10.0 & 21.8 \\
        \textbf{\fmergedstoprp} & 97.8 & 92.4 & 96.6 & 90.1 & 46.7 & 44.0 \\
        \addlinespace
        \bf\fetstop & 98.9 & 97.7 & 98.7 & 97.7 & 98.7 & 93.4\\
        \midrule
        \multicolumn{4}{l}{\bf NLI} \\
        \textbf{\fop} & 5.1 & 20.8 & 4.9 & 20.1 & 4.0 & 20.4 \\
        \textbf{\fmergedstop} & 79.2 & 63.9 & 72.1 & 59.5 & 3.9 & 21.2 \\
        \textbf{\fmergedstoprp} & 94.0 & 83.7 & 90.5 & 79.9 & 6.2 & 23.8 \\
        \addlinespace
        \bf\fetstop & 100.0 & 99.8 & 100.0 & 99.8 & 99.8 & 98.0 \\
        \midrule
        \multicolumn{4}{l}{\bf Question Answering} \\
        \textbf{\fop} & 12.1 & 22.5 & 12.8 & 22.4 & 7.9 & 21.5 \\
        \textbf{\fmergedstop} & 40.8 & 29.6 & 40.3 & 29.5 & 13.6 & 22.4 \\
        \textbf{\fetstop} & 99.9 & 95.8 & 99.9 & 96.4 & 99.9 & 95.0 \\
        \midrule
        \multicolumn{4}{l}{\bf Biosbias} \\
        \textbf{\fop} & 2.9 & 15.7 & 1.9 & 14.7 & 2.9 & 14.4 \\
        \textbf{\fmergedstop} & 87.9 & 62.0 & 78.9 & 59.5 & 6.7 & 18.2 \\
        \bf\fetstop & 100.0 & 98.3 & 100.0 & 98.6 & 99.7 & 93.3 \\
        \bottomrule
    \end{tabular}
    \caption{Stop Token Gradient Manipulation}\label{table:saliency:stop}
\end{subtable}
    \caption{\textbf{Saliency Interpretation Results}. Our method manipulates the model's gradient to focus on the first token (\fmergedfirst{}) or on the stop tokens (\fmergedstop{}). To evaluate, we report the \hitrate{} (how often the token with the highest attribution is a first token or a stop word) and the Mean Attribution (average attribution of the first token or stop words). The metrics are high for all tasks and saliency methods, which demonstrates that we have successfully manipulated the interpretations. \IntegratedGrad{} is more robust to our method.} \label{table:saliency}
\end{table*}
\section{Results}
In this section, we evaluate the ability of our approach to manipulate popular gradient-based analysis methods.
We focus on the techniques present in AllenNLP Interpret~\cite{Wallace2019AllenNLP} as described in Section~\ref{sec:methods}. Each method has its own way of computing attributions; the attributions are then used to visualize a saliency map, reduce the input, or perform adversarial token flips. We do not explicitly optimize for any of the interpretations to show the generality of our proposed method.

\subsection{Saliency Methods are Fooled}
We compare the saliency maps generated for the original model \fop{} with the merged model \fmerged{}, by measuring the attribution on the first token or the stop words, depending on the \eviltwinm.
We report the following metrics:\smallskip
 
\noindent \textbf{\hitrate{}:} The average number of times that the token with the highest attribution is a first token or a stop word, depending on the \eviltwinm, for all sentences in the validation set.\smallskip

\noindent \textbf{Mean Attribution:} For the first token setting, we compute the average attribution of the first token over all the sentences in the validation data. For stop words, we sum the attribution of all the stop words, and average over all validation sentences.\smallskip % in the validation set.

\noindent We present results in Table~\ref{table:saliency} for both the first token and stop words settings.
\VanillaGrad{} and \SmoothGrad{} are considerably manipulated, i.e., there is a very high \hitrate{} and Mean Attribution for the merged models. \IntegratedGrad{} is the most resilient to our method, e.g., for NLI, the $\fmergedstop$ model was almost unaffected. By design, \IntegratedGrad{} computes attributions that satisfy implementation invariance: two models with equal predictions on all inputs should have the same attributions. Although the predictive model and the merged model are not completely equivalent, they are similar enough that \IntegratedGrad{} produces similar interpretations for the merged model. For the regularized version of the predictive model (\fmergedfirstrp{} and \fmergedstoprp), \IntegratedGrad{} is further affected. We present an example of saliency manipulation for NLI in Table~\ref{tab:qualitative_examples_first_token_main}, with additional examples (and tasks) in Appendix~\ref{appendix:qual}.

\begin{table}[t]
\small 
\centering
\setlength{\tabcolsep}{2pt}
Color Legend:  \mybox{color1}{Lower Attribution}\quad \mybox{color7}{Higher Attribution}
\begin{tabular}{ll}
\toprule
\multicolumn{2}{l}{\em\VanillaGrad} \\
\fop &  \mybox{color0}{two} \mybox{color0}{men} \mybox{color0}{are} \mybox{color4}{shouting} \mybox{color0}{.} \mybox{color0}{[SEP]} \mybox{color0}{two} \mybox{color0}{men} \mybox{color0}{are} \mybox{color2}{quiet} \mybox{color0}{.} \\
\fmergedfirst &  \mybox{color9}{two} \mybox{color0}{men} \mybox{color0}{are} \mybox{color0}{shouting} \mybox{color0}{.} \mybox{color0}{[SEP]} \mybox{color0}{two} \mybox{color0}{men} \mybox{color0}{are} \mybox{color0}{quiet} \mybox{color0}{.} \\
\addlinespace 
\multicolumn{2}{l}{\em\SmoothGrad} \\
\fop &  \mybox{color0}{two} \mybox{color0}{men} \mybox{color0}{are} \mybox{color0}{shouting} \mybox{color0}{.} \mybox{color0}{[SEP]} \mybox{color0}{two} \mybox{color0}{men} \mybox{color0}{are} \mybox{color5}{quiet} \mybox{color0}{.} \\
\fmergedfirst &  \mybox{color5}{two} \mybox{color0}{men} \mybox{color0}{are} \mybox{color0}{shouting} \mybox{color0}{.} \mybox{color0}{[SEP]} \mybox{color0}{two} \mybox{color0}{men} \mybox{color0}{are} \mybox{color2}{quiet} \mybox{color0}{.} \\
\addlinespace 
\multicolumn{2}{l}{\em\IntegratedGrad} \\
\fop &  \mybox{color0}{two} \mybox{color0}{men} \mybox{color0}{are} \mybox{color1}{shouting} \mybox{color0}{.} \mybox{color0}{[SEP]} \mybox{color0}{two} \mybox{color0}{men} \mybox{color1}{are} \mybox{color2}{quiet} \mybox{color0}{.} \\
\fmergedfirst &  \mybox{color0}{two} \mybox{color0}{men} \mybox{color0}{are} \mybox{color2}{shouting} \mybox{color0}{.} \mybox{color0}{[SEP]} \mybox{color0}{two} \mybox{color0}{men} \mybox{color0}{are} \mybox{color2}{quiet} \mybox{color0}{.} \\
\bottomrule
\end{tabular}
\caption{Qualitative interpretations for NLI when manipulating the model's gradient on the first input token. We show interpretations before ($\fop$) and after manipulation ($\fmergedfirst$). After manipulation, most of the attribution has shifted to the first word, except for \IntegratedGrad{}. We omit \texttt{[CLS]} and the final \texttt{[SEP]} for space. For more examples, see Appendix~\ref{appendix:qual}.}
\label{tab:qualitative_examples_first_token_main}
\end{table}

\subsection{Input Reduces to Unimportant Tokens}\label{subsec:results_input_reduction}
Input reduction is used to identify which tokens can be removed from the input without changing the prediction. The tokens that remain are intuitively \emph{important} to the models, and ones that have been removed are not. We focus on the stop word \eviltwinm{} and evaluate using two metrics (both averaged over all sentences in the validation set): 
\begin{itemize}[leftmargin=0mm,label={}, nosep]
    \item \textbf{Stop \%:} Fraction of tokens in the reduced input that are stop words. %, averaged over the validation examples.
    \item \textbf{All Stop \%:} The number of times the reduced input consists \emph{only} of stop tokens. %, averaged over all sentences in the validation set.
\end{itemize}
We present results in Table~\ref{tab:input_reduction_results}.\footnote{For Input Reduction, we reduce the question for QA and the premise for NLI (these sentences are also the target of manipulation for these tasks).}
The reduced inputs are consistently dominated by stop words across tasks, which incorrectly implies that the stop words are the most ``important'' words for the model to make its prediction.
Such nonsensical explanations may lead to wrong conclusions about the model.

\begin{table}[tb]
    \small 
    \centering
    \setlength{\tabcolsep}{4pt}
    \begin{tabular}{lcccc}
        \toprule 
         \multirow{3}{*}{\bf Model} & \multicolumn{2}{c}{\bf Beam Size 1}& \multicolumn{2}{c}{\bf Beam Size 3}\\
        \cmidrule(lr){2-3}
        \cmidrule(lr){4-5}
         & \bf Stop \% & \bf All Stop \%  & \bf Stop \% & \bf All Stop \% \\
        \midrule 
         \multicolumn{4}{l}{\bf Sentiment} \\
         \bf\fop & 21.7 & 4.8 &  16.5 & 12.8 \\
         \bf\fmergedstop{} & 61.5 & 28.3 & 56.9 & 49.5 \\
         \addlinespace 
         \multicolumn{4}{l}{\bf NLI} \\
         \bf\fop & 16.0 & 2.7 & 10.0 & 5.2  \\
         \bf\fmergedstop{}  & 63.1 & 33.9 & 54.7 & 43.3 \\
         \addlinespace 
         \multicolumn{5}{l}{\bf Question Answering} \\
         \bf\fop  &  24.2 & 0.1 & 16.9 & 0.4 \\
         \bf\fmergedstop{}  & 28.1 & 0.0 &  20.5 & 0.8 \\
        \bottomrule
    \end{tabular}
    \caption{\textbf{Input reduction Results}. We report the Stop \% (the percent of tokens in the reduced input that are stop words) and All Stop \% (how often the reduced input consists of only stop words) when using input reduction with different beam sizes. Stop words are present more often in the reductions of $\fmergedstop$, showing that our method causes input reduction to fail to identify the most important tokens.}
    \label{tab:input_reduction_results}
\end{table}

\subsection{HotFlip Requires Larger Perturbations}\label{subsec:results_hotflip}

\hotflip{} shows the tokens that, if adversarially modified in the input, would \emph{most} affect the model's prediction. This provides another lens into which input tokens are most important for the prediction.
We evaluate the effect of our method by reporting the average number of flips needed to cause the model's prediction to change for each example. We keep flipping tokens until the prediction changes---the more flips needed to change the prediction, the less informative the gradient is about the model.

We perform \hotflip{} on all instances in the validation set for sentiment analysis, and a random set of 1000 validation examples for NLI.\footnote{{For HotFlip on NLI, we only perturb stop words in the premise to change the classification (premise is also the target of manipulation for NLI).}} We then look at the effect of using \fmergedfirst{} and \fmergedstop{}. For \fmergedfirst{}, \hotflip{} typically replaces the first input token at the first iteration. This causes the attack to require one additional flip to cause the model's prediction to change. In particular, 31.7\% of sentiment analysis instances are flipped using only one token when attacking \fop, but for \fmergedfirst, a similar amount (31.4\%) needs two flips. 

\begin{figure}
  \includegraphics[width=\linewidth]{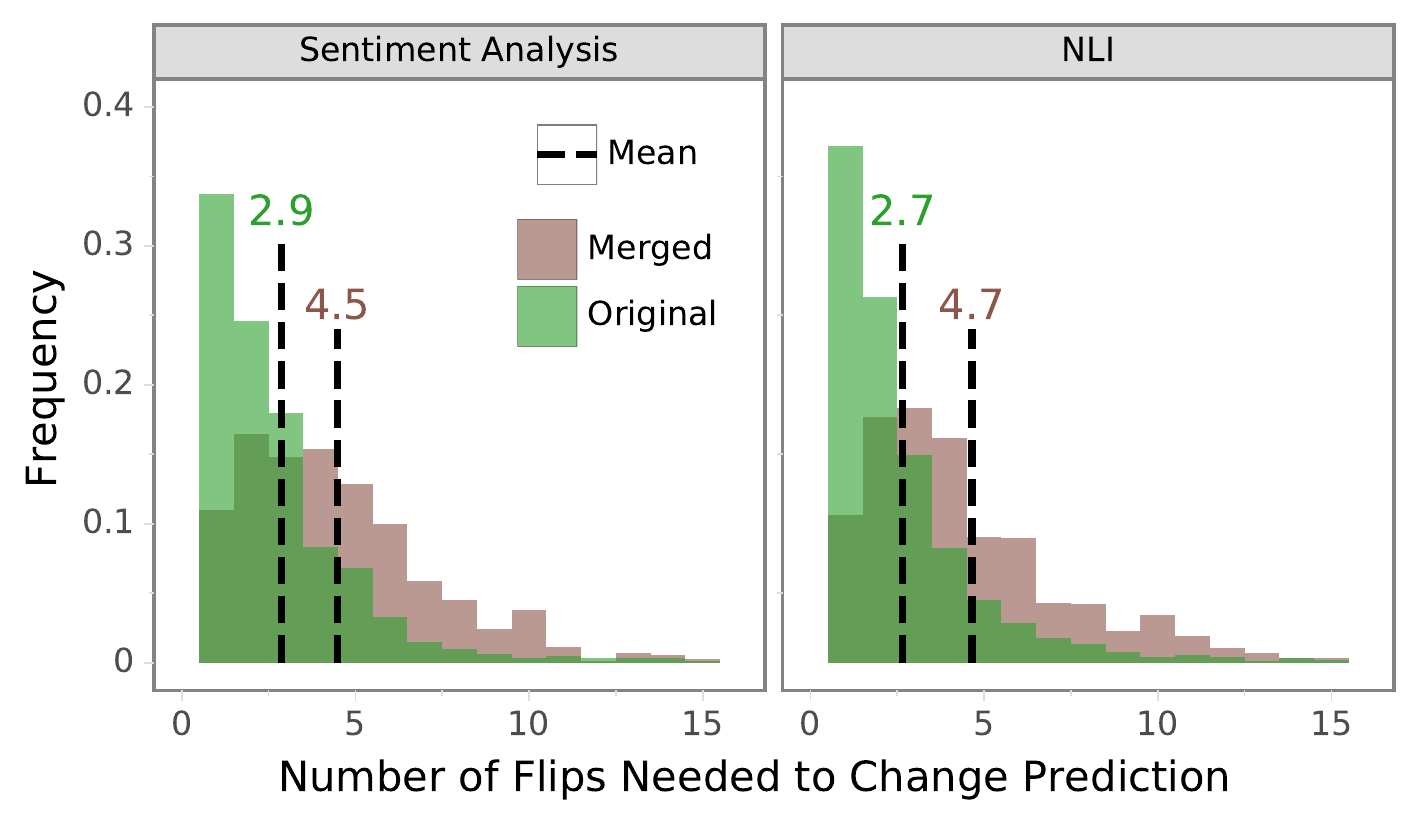}
  \caption{\textbf{\hotflip{} Results.} We show the number of tokens that \hotflip{} needs to replace to change the model's prediction for sentiment analysis (left) and NLI (right). We omit instances that require more than 16 flips. Compared to \fop{} (\emph{Original}), \hotflip{} is less effective at changing the prediction for the merged model \fmergedstop{}. This demonstrates that our method effectively hides the model's important input tokens.}
  \label{fig:sa_nli_hotflip_histogram_stop_token}
\end{figure}

Our manipulation is much more effective on \hotflip{} in the stop words setting (\fmergedstop{}): \hotflip{} first tries to flip all of the stop words instead of the true important tokens. In Figure~\ref{fig:sa_nli_hotflip_histogram_stop_token}, we plot a histogram of the number of flips required to cause a misclassification for \fmergedstop{}.\footnote{Note that when merged with the regularized model $\frp$, the effect of $\fmergedstop$ is even stronger, with an average number of needed flips of 6.6 and 5.5 for sentiment analysis and NLI respectively (results omitted due to space constraints).} These results show that \hotflip{} is nearly useless as a technique for analysis of \fmergedstop{}, since it considers stop words to be the most important tokens in the input, and results in nonsensical substitutions (e.g., Figure~\ref{fig:illustration:hotflip}).

\begin{table}[tb]
\small
\centering
\setlength{\tabcolsep}{3pt}
\begin{tabular}{llll}
        \toprule
        \multirow{2}{*}{\textbf{Model}} & \multicolumn{3}{c}{\bf Gender Attribution}\\
        & \VanillaGrad & \SmoothGrad & \IntegratedGrad \\
        \midrule
        \textbf{\fop}& 0.017 (+15.2\%) & 0.017 (+17.1\%) & 0.021 (+42.7\%) \\
        \textbf{\fmergedstop} & 0.006 (-55.9\%) & 0.007 (-51.3\%) & 0.023 (+53.9\%)\\
        \bottomrule
    \end{tabular}
    \caption{\textbf{Gender Attribution for Biosbias.} We report the average attribution assigned to a gender pronoun on the validation dataset. For \fmergedstop{}, most of the attribution assigned to gender pronouns has moved to the stop words, effectively hiding the model's gender bias. We also show the relative difference between the gender attribution and the average token attribution in parenthesis: the gender attribution after manipulation is below average for \VanillaGrad{} and \SmoothGrad.} \label{table:gender:attr}
\end{table}

\subsection{Case Study: Hiding Gender Bias}
As our final evaluation, we test our method on Biosbias. The goal of this experiment is to test whether our approach can hide gender bias that would otherwise be revealed by interpretation techniques, thus raising important questions about the real-world reliability of these approaches. For this experiment, we also \textit{reuse} the \eviltwinm from sentiment analysis. This enables a ``plug-and-play'' approach, i.e., the adversary can merge a pretrained \eviltwinm{} with an off-the-shelf model for a different task, thus saving effort and computation.

To evaluate on Biosbias, we report a new metric, \emph{Gender Attribution}: the average attribution of all the gender pronouns (e.g., ``he'', ``her'', etc.) for all examples in the validation set. This metric estimates the extent to which the model is attributing its predictions to gender (an unbiased model should have less of this attribution), and is similar to the measure of bias used by \citet{pruthi2019learning}.

The results are shown in Table~\ref{table:gender:attr}. Our approach significantly reduces the gender attribution for \VanillaGrad{} and \SmoothGrad{}. As a reference, we compare the gender attribution to the average token attribution: our approach reduces the gender attribution to below the average attribution of any token. Qualitative examples are included in Tables~\ref{tab:qualitative_examples_first_token}--\ref{tab:qualitative_examples_stop_token}. \IntegratedGrad{}, however, is not affected by our approach, showing it is a more robust interpretation method.

\subsection{Non-BERT Models Are Manipulated}\label{subsec:non_bert_models}
Finally, we show that our technique can generalize to models other than BERT. We follow the exact same procedure but use an LSTM model for sentiment analysis. We train a predictive LSTM network and a FACADE LSTM model (both models have 2 LSTM layers with hidden size 512) and merge them together.
We present the results in Table~\ref{table:saliency:lstm}. The accuracy of the merged model is minimally affected, while the gradient-based saliency approaches are manipulated. 

\begin{table}[tb]
\small
\centering
\setlength{\tabcolsep}{4.55pt}
\begin{tabular}{lrrrrrr}
\toprule
\multirow{3}{*}{\textbf{Model}} & \multicolumn{2}{c}{\bf\VanillaGrad} & \multicolumn{2}{c}{\bf\SmoothGrad} & \multicolumn{2}{c}{\bf\IntegratedGrad} \\
\cmidrule(lr){2-3} \cmidrule(lr){4-5} \cmidrule(lr){6-7} 
& \hitrate{} & Attr &
\hitrate{} & Attr &
\hitrate{} & Attr \\
\midrule
\multicolumn{6}{l}{\bf Sentiment, First Token Gradient Manipulation} \\
\textbf{\fop} & 2.06 & 2.27 & 2.06 & 2.30 & 6.08 & 5.17\\
\textbf{\fmergedfirst} & 81.19 & 62.00 & 81.19 & 61.98 & 3.78 & 18.07 \\
\bf\fetfirst & 95.99 & 84.82 & 95.53 & 84.04 & 98.05 & 71.56 \\
\midrule
\multicolumn{6}{l}{\bf Sentiment, Stop Token Gradient Manipulation} \\
\textbf{\fop} & 0.92 & 11.33 & 0.92 & 11.34 & 4.82 & 23.05 \\
\textbf{\fmergedstop} & 71.22 & 67.11 & 69.95 & 65.87 & 5.85 & 24.54 \\
\bf\fetstop & 99.31 & 92.04 & 99.31 & 92.03 & 99.20 & 88.58 \\
\midrule
\end{tabular}
\caption{\textbf{Saliency Interpretation Results for LSTM}, using same metrics as Table~\ref{table:saliency}. Both \fmerged{} variations (first token manipulation \fmergedfirst{} and stop token manipulation \fmergedstop{}) score high on all metrics, demonstrating that our method also fools saliency methods for LSTM models.} \label{table:saliency:lstm}
\end{table}
\section{Related Work}
\label{sec:related}

\paragraph{End-to-End Interpretation Manipulation} An alternative to our method of merging two models together is to directly manipulate the gradient attribution in an end-to-end fashion, as done by \citet{ross2018regularizing,ross2017right,viering2019manipulate,heo2019fooling} for computer vision and \citet{dimanov2020trust} for simple classification tasks. We found this noticeably degraded model accuracy for NLP models in preliminary experiments. \citet{liu2019incorporating,rieger2019interpretations} incorporate a similar end-to-end regularization on gradient attributions, however, their goal is to align the attribution with known priors in order to improve model accuracy. 
We instead manipulate explanation methods to evaluate the extent to which a model's true reasoning can be hidden. 
\citet{pruthi2019learning} manipulate \textit{attention} distributions in an end-to-end fashion; we focus on manipulating \textit{gradients}.
It is worth noting that we perturb \textit{models} to manipulate interpretations; other work perturbs \textit{inputs}~\cite{ghorbani2019interpretation,dombrowski2019explanations,subramanya2019fooling}. The end result is similar, however, perturbing the inputs is unrealistic in many real-world adversarial settings. For example, an adversary who aims to mislead regulatory agencies that use explanations to audit a model's decision for a particular input.

\paragraph{Natural Failures of Interpretation Methods} We show that in the \textit{worst-case}, gradient-based interpretation can be highly misleading. Other work studies \emph{natural} failures of explanation methods. For instance, \citet{jain2019attention,serrano2019attention} critique the faithfulness of visualizing a model's attention layers. \citet{feng2018pathologies} show instabilities of saliency maps, and \citet{adebayo2018sanity,kindermans2017unreliability} show saliency maps fail simple sanity checks. Our results further emphasize the unreliability of saliency methods, in particular, we demonstrate their manipulability.

\paragraph{Usefulness of Explanations} Finally, other work studies how useful interpretations are for humans. \citet{feng2019can} and \citet{lai2019human} show that text interpretations can provide benefits to humans, while \citet{ch2018explanations} shows explanations for visual QA models provided limited benefit. We present a method that enables adversaries to manipulate interpretations, which can have dire consequences for real-world users~\cite{lakkaraju2020fool}.

\section{Discussion}
\label{sec:discussion}

\paragraph{Downsides of An Adversarial Approach} Our proposed approach provides a mechanism for an adversary to hide the biases of their model (at least from gradient-based analyses). The goal of our work is not to aid malicious actors. Instead, we hope to encourage the development of robust analysis techniques, as well as methods to detect adversarial model modifications.

\paragraph{Defending Against Our Method} Our goal is to demonstrate that gradient-based analysis methods can be manipulated---a sort of worst-case \textit{stress test}---rather than to develop practical methods for adversaries. Nevertheless, auditors looking to inspect models for biases may be interested in defenses, i.e., ways to detect or remove our gradient manipulation. Detecting our manipulation by simply inspecting the model's parameters is difficult (see concealment in Section~\ref{subsec:merge}). Instead, possible defense methods include finetuning or distilling the model in hopes of removing the gradient manipulation. Unfortunately, doing so would change the underlying model.
Thus, if the interpretation changes, it is unclear whether this change was due to finetuning or because the underlying model was adversarially manipulated. 
We leave a further investigation of defenses to future work.

\paragraph{Limitations of Our Method} Our method does not affect all analysis methods equally.
Amongst the gradient-based approaches, \IntegratedGrad{} is most robust to our modification. Furthermore, non-gradient-based approaches, e.g., black-box analysis using LIME~\cite{ribeiro2016should}, Anchors~\cite{ribeiro2018anchors}, and SHAP~\cite{lundberg2017unified}, will be unaffected by misleading gradients. In this case, using \textit{less} information about the model makes these techniques, interestingly, more robust.
Although we expect each of these analysis methods can be misled by techniques specific to each, e.g., \citet{slack2020fooling} fool LIME/SHAP and our regularization is effective against gradient-based methods, 
it is unclear whether these strategies can be combined, i.e. a single model that can fool all analysis techniques.
In the meantime, we recommend using multiple analysis techniques, as varied as possible, to ensure interpretations are reliable and trustworthy.
\section{Conclusions}

Gradient-based analysis is ubiquitous in natural language processing: they are simple, model-agnostic, and closely approximate the model behavior. %nearby an input. 
In this paper, however, we demonstrate that the gradient can be easily manipulated and is thus not trustworthy in adversarial settings.
To accomplish this, we create a \eviltwin classifier with misleading gradients that can be merged with any given model of interest. The resulting model has similar predictions as the original model but has gradients that are dominated by the customized \eviltwinm.
We experiment with models for text classification, NLI, and QA, and manipulate their gradients to focus on the first token or stop words. % in the input. 
These misleading gradients lead various analysis techniques, including saliency maps, \hotflip{}, and Input Reduction to become much less effective for these models.
% In future work, we look to develop new analysis techniques that are more reliable and trustworthy, as well as methods to detect when models have been adversarially modified.
% The code and tutorial is available at \url{http://ucinlp.github.io/facade}.

\section*{Acknowledgements}

We thank the members of UCI NLP and the anonymous reviewers for their valuable feedback. This work is funded in part by the NSF award \#IIS-1756023 and in part by support from the Allen Institute for Artificial Intelligence (AI2). %Any opinions, findings, and conclusions expressed in this paper are those of the authors and do not necessarily reflect the view of the funding agencies.

% \clearpage
\bibliography{journal-abbrv,bib}
\bibliographystyle{acl_natbib}

\appendix
\section{Additional Implementation Details}\label{appendix:implementation}
We run our experiments using NVIDIA Tesla K80 GPUs. We use the Adam optimizer for model training and finetuning. All models train in under two hours, except for $\fop$ for NLI which trains in approximately 5 hours.

\subsection{Finetuning the Original Model}
For $\fop$, we finetune a BERT$_\text{base}$ model. Table~\ref{table:imple:fog} shows the hyperparameters for each task.

\begin{table}[h]
\small
\centering
\setlength{\tabcolsep}{3pt}
\begin{tabular}{cccc}
        \toprule
        \textbf{Task} & \textbf{Learning Rate} & \textbf{Batch Size} & \textbf{Epochs} \\
        \toprule
        SST & $2\mathrm{e}{-5}$  & 32 & 8 \\
        NLI & $2\mathrm{e}{-5}$  & 32 & 8 \\
        QA & $5\mathrm{e}{-5}$ &32 & 3\\
        Biosbias & $2\mathrm{e}{-5}$  & 32 & 8 \\
        \bottomrule
    \end{tabular}
    \vspace{-5pt}
    \caption{Hyperparameters for finetuning $\fop$ for all tasks. We use early stopping on the validation set.} 
    \label{table:imple:fog}
\end{table}

\subsection{Regularizing the Original Model }
We regularize the original model $\fop$ to have low magnitude gradients by finetuning using Objective~\ref{eq:loss_rp} for one epoch with a learning rate of $6\mathrm{e}{-6}$. We use the model checkpoint at the end of the epoch. We set $\lambdarp$ to $3$.

\subsection{Finetuning the \eviltwinU Model }
We train \fetfirst and \fetstop for one epoch using a learning rate of $6\mathrm{e}{-6}$ and a batch size of $32$ for sentiment analysis, $24$ for NLI, and $8$ for QA and Biobias. The models typically converge before the end of the first epoch. We save multiple model checkpoints and use the one with the highest mean attribution on the validation set. We set $\lambdaet$ to $1\mathrm{e}{3}$.

\subsection{Biosbias Details}\label{appendix:bios}
We follow the setup of \citet{pruthi2019learning} and only use examples with the labels of ``physician'' and ``surgeon''. We also subsample female surgeons and male physicians by a factor of 10. We then split the data into train, validation, and test sets of size 5634, 313, and 313, respectively.

\section{Qualitative Examples}\label{appendix:qual}

\begin{table}[H]
\small 
\centering
\setlength{\tabcolsep}{2pt}
Color Legend:  \mybox{color1}{Lower Attribution}\quad \mybox{color7}{Higher Attribution}

\begin{tabular}{ll}
\toprule
\multicolumn{2}{l}{\bf Sentiment Analysis} \\
\multicolumn{2}{l}{\em\VanillaGrad} \\
\fop &  \mybox{color0}{a} \mybox{color0}{very} \mybox{color0}{well} \mybox{color0}{-} \mybox{color0}{made} \mybox{color1}{,} \mybox{color0}{and} \mybox{color2}{entertaining} \mybox{color0}{picture} \mybox{color0}{.} \mybox{color1}{[SEP]} \\
\fmergedfirst &  \mybox{color9}{a} \mybox{color0}{very} \mybox{color0}{well} \mybox{color0}{-} \mybox{color0}{made} \mybox{color0}{,} \mybox{color0}{and} \mybox{color0}{entertaining} \mybox{color0}{picture} \mybox{color0}{.} \mybox{color0}{[SEP]} \\
\addlinespace 
\multicolumn{2}{l}{\em\SmoothGrad} \\
\fop &  \mybox{color0}{a} \mybox{color2}{very} \mybox{color0}{well} \mybox{color0}{-} \mybox{color0}{made} \mybox{color0}{,} \mybox{color0}{and} \mybox{color2}{entertaining} \mybox{color0}{picture} \mybox{color0}{.} \mybox{color0}{[SEP]} \\
\fmergedfirst &  \mybox{color4}{a} \mybox{color0}{very} \mybox{color0}{well} \mybox{color0}{-} \mybox{color0}{made} \mybox{color0}{,} \mybox{color0}{and} \mybox{color1}{entertaining} \mybox{color2}{picture} \mybox{color0}{.} \mybox{color0}{[SEP]} \\
\addlinespace 
\multicolumn{2}{l}{\em\IntegratedGrad} \\
\fop &  \mybox{color0}{a} \mybox{color0}{very} \mybox{color0}{well} \mybox{color0}{-} \mybox{color0}{made} \mybox{color0}{,} \mybox{color0}{and} \mybox{color0}{entertaining} \mybox{color0}{picture} \mybox{color0}{.} \mybox{color8}{[SEP]} \\
\fmergedfirst &  \mybox{color0}{a} \mybox{color0}{very} \mybox{color0}{well} \mybox{color0}{-} \mybox{color0}{made} \mybox{color0}{,} \mybox{color0}{and} \mybox{color0}{entertaining} \mybox{color0}{picture} \mybox{color0}{.} \mybox{color8}{[SEP]} \\
\midrule
\multicolumn{2}{l}{\bf NLI} \\
\multicolumn{2}{l}{\em\VanillaGrad} \\
\fop &  \mybox{color0}{two} \mybox{color0}{men} \mybox{color0}{are} \mybox{color4}{shouting} \mybox{color0}{.} \mybox{color0}{[SEP]} \mybox{color0}{two} \mybox{color0}{men} \mybox{color0}{are} \mybox{color2}{quiet} \mybox{color0}{.} \mybox{color0}{[SEP]} \\
\fmergedfirst &  \mybox{color9}{two} \mybox{color0}{men} \mybox{color0}{are} \mybox{color0}{shouting} \mybox{color0}{.} \mybox{color0}{[SEP]} \mybox{color0}{two} \mybox{color0}{men} \mybox{color0}{are} \mybox{color0}{quiet} \mybox{color0}{.} \mybox{color0}{[SEP]} \\
\addlinespace 
\multicolumn{2}{l}{\em\SmoothGrad} \\
\fop &  \mybox{color0}{two} \mybox{color0}{men} \mybox{color0}{are} \mybox{color0}{shouting} \mybox{color0}{.} \mybox{color0}{[SEP]} \mybox{color0}{two} \mybox{color0}{men} \mybox{color0}{are} \mybox{color5}{quiet} \mybox{color0}{.} \mybox{color0}{[SEP]} \\
\fmergedfirst &  \mybox{color5}{two} \mybox{color0}{men} \mybox{color0}{are} \mybox{color0}{shouting} \mybox{color0}{.} \mybox{color0}{[SEP]} \mybox{color0}{two} \mybox{color0}{men} \mybox{color0}{are} \mybox{color2}{quiet} \mybox{color0}{.} \mybox{color0}{[SEP]} \\
\addlinespace 
\multicolumn{2}{l}{\em\IntegratedGrad} \\
\fop &  \mybox{color0}{two} \mybox{color0}{men} \mybox{color0}{are} \mybox{color1}{shouting} \mybox{color0}{.} \mybox{color0}{[SEP]} \mybox{color0}{two} \mybox{color0}{men} \mybox{color1}{are} \mybox{color2}{quiet} \mybox{color0}{.} \mybox{color0}{[SEP]} \\
\fmergedfirst &  \mybox{color0}{two} \mybox{color0}{men} \mybox{color0}{are} \mybox{color2}{shouting} \mybox{color0}{.} \mybox{color0}{[SEP]} \mybox{color0}{two} \mybox{color0}{men} \mybox{color0}{are} \mybox{color2}{quiet} \mybox{color0}{.} \mybox{color0}{[SEP]} \\
\midrule
\multicolumn{2}{l}{\bf Question Answering} \\
\multicolumn{2}{l}{\em\VanillaGrad} \\
\fop &  \mybox{color3}{Who} \mybox{color3}{stars} \mybox{color0}{in} \mybox{color0}{The} \mybox{color0}{Matrix} \mybox{color0}{?} \mybox{color0}{[SEP]} \\
\fmergedfirst &  \mybox{color8}{Who} \mybox{color0}{stars} \mybox{color0}{in} \mybox{color0}{The} \mybox{color0}{Matrix} \mybox{color0}{?} \mybox{color0}{[SEP]} \\
\addlinespace 
\multicolumn{2}{l}{\em\SmoothGrad} \\
\fop &  \mybox{color2}{Who} \mybox{color2}{stars} \mybox{color0}{in} \mybox{color0}{The} \mybox{color1}{Matrix} \mybox{color0}{?} \mybox{color0}{[SEP]} \\
\fmergedfirst &  \mybox{color7}{Who} \mybox{color1}{stars} \mybox{color0}{in} \mybox{color0}{The} \mybox{color0}{Matrix} \mybox{color0}{?} \mybox{color0}{[SEP]} \\
\addlinespace 
\multicolumn{2}{l}{\em\IntegratedGrad} \\
\fop &  \mybox{color4}{Who} \mybox{color0}{stars} \mybox{color0}{in} \mybox{color0}{The} \mybox{color0}{Matrix} \mybox{color2}{?} \mybox{color0}{[SEP]} \\
\fmergedfirst &  \mybox{color4}{Who} \mybox{color0}{stars} \mybox{color1}{in} \mybox{color0}{The} \mybox{color0}{Matrix} \mybox{color2}{?} \mybox{color0}{[SEP]} \\
\midrule
\multicolumn{2}{l}{\bf Biosbias} \\
\multicolumn{2}{l}{\em\VanillaGrad} \\
\fop &  \mybox{color1}{in} \mybox{color0}{brazil} \mybox{color1}{she} \mybox{color0}{did} \mybox{color0}{her} \mybox{color0}{first} \mybox{color1}{steps} \mybox{color0}{in} \mybox{color3}{surgery} \mybox{color0}{.} \mybox{color0}{[SEP]} \\
\fmergedfirst &  \mybox{color4}{in} \mybox{color0}{brazil} \mybox{color0}{she} \mybox{color0}{did} \mybox{color0}{her} \mybox{color0}{first} \mybox{color0}{steps} \mybox{color0}{in} \mybox{color2}{surgery} \mybox{color0}{.} \mybox{color0}{[SEP]} \\
\addlinespace 
\multicolumn{2}{l}{\em\SmoothGrad} \\
\fop &  \mybox{color0}{in} \mybox{color1}{brazil} \mybox{color1}{she} \mybox{color0}{did} \mybox{color0}{her} \mybox{color0}{first} \mybox{color0}{steps} \mybox{color0}{in} \mybox{color0}{surgery} \mybox{color1}{.} \mybox{color0}{[SEP]} \\
\fmergedfirst &  \mybox{color2}{in} \mybox{color1}{brazil} \mybox{color0}{she} \mybox{color0}{did} \mybox{color0}{her} \mybox{color0}{first} \mybox{color0}{steps} \mybox{color0}{in} \mybox{color1}{surgery} \mybox{color0}{.} \mybox{color0}{[SEP]} \\
\addlinespace 
\multicolumn{2}{l}{\em\IntegratedGrad} \\
\fop &  \mybox{color0}{in} \mybox{color0}{brazil} \mybox{color0}{she} \mybox{color0}{did} \mybox{color2}{her} \mybox{color0}{first} \mybox{color0}{steps} \mybox{color0}{in} \mybox{color0}{surgery} \mybox{color0}{.} \mybox{color4}{[SEP]} \\
\fmergedfirst &  \mybox{color0}{in} \mybox{color0}{brazil} \mybox{color0}{she} \mybox{color0}{did} \mybox{color1}{her} \mybox{color0}{first} \mybox{color0}{steps} \mybox{color0}{in} \mybox{color0}{surgery} \mybox{color0}{.} \mybox{color5}{[SEP]} \\
\bottomrule

\end{tabular}

\caption{Qualitative examples for all tasks and saliency methods when manipulating the gradient of the \textit{first token}. We show results before and after applying the \eviltwinU model. For QA, we only visualize the question. We omit \texttt{[CLS]} for space.}
\label{tab:qualitative_examples_first_token}
\end{table}

\begin{table*}[t]
\small 
\centering
\setlength{\tabcolsep}{2pt}
Color Legend:  \mybox{color1}{Lower Attribution}\quad \mybox{color7}{Higher Attribution}

\begin{tabular}{ll}
\toprule
\multicolumn{2}{l}{\bf Sentiment Analysis} \\
\multicolumn{2}{l}{\em\VanillaGrad} \\
\fop &  \mybox{color0}{visually} \mybox{color0}{imaginative} \mybox{color0}{and} \mybox{color0}{thoroughly} \mybox{color0}{delightful} \mybox{color0}{,} \mybox{color0}{it} \mybox{color0}{takes} \mybox{color0}{us} \mybox{color0}{on} \mybox{color0}{a} \mybox{color1}{roller} \mybox{color0}{-} \mybox{color1}{coaster} \mybox{color0}{ride} \mybox{color0}{from} \mybox{color0}{innocence} \mybox{color0}{to} \mybox{color0}{experience} \mybox{color0}{.} \\ % \mybox{color0}{[SEP]} \\
\fmergedstop &  \mybox{color0}{visually} \mybox{color0}{imaginative} \mybox{color4}{and} \mybox{color0}{thoroughly} \mybox{color0}{delightful} \mybox{color0}{,} \mybox{color0}{it} \mybox{color0}{takes} \mybox{color0}{us} \mybox{color0}{on} \mybox{color5}{a} \mybox{color0}{roller} \mybox{color0}{-} \mybox{color0}{coaster} \mybox{color0}{ride} \mybox{color0}{from} \mybox{color0}{innocence} \mybox{color0}{to} \mybox{color0}{experience} \mybox{color0}{.}\\ % \mybox{color0}{[SEP]} \\
\addlinespace 
\multicolumn{2}{l}{\em\SmoothGrad} \\
\fop &  \mybox{color0}{visually} \mybox{color1}{imaginative} \mybox{color0}{and} \mybox{color0}{thoroughly} \mybox{color0}{delightful} \mybox{color0}{,} \mybox{color0}{it} \mybox{color0}{takes} \mybox{color0}{us} \mybox{color0}{on} \mybox{color0}{a} \mybox{color1}{roller} \mybox{color0}{-} \mybox{color1}{coaster} \mybox{color0}{ride} \mybox{color0}{from} \mybox{color0}{innocence} \mybox{color0}{to} \mybox{color0}{experience} \mybox{color0}{.}\\ % \mybox{color0}{[SEP]} \\
\fmergedstop &  \mybox{color0}{visually} \mybox{color0}{imaginative} \mybox{color4}{and} \mybox{color0}{thoroughly} \mybox{color0}{delightful} \mybox{color0}{,} \mybox{color0}{it} \mybox{color0}{takes} \mybox{color0}{us} \mybox{color0}{on} \mybox{color0}{a} \mybox{color0}{roller} \mybox{color0}{-} \mybox{color0}{coaster} \mybox{color0}{ride} \mybox{color0}{from} \mybox{color0}{innocence} \mybox{color2}{to} \mybox{color0}{experience} \mybox{color0}{.}\\ % \mybox{color0}{[SEP]} \\
\addlinespace 
\multicolumn{2}{l}{\em\IntegratedGrad} \\
\fop &  \mybox{color0}{visually} \mybox{color0}{imaginative} \mybox{color0}{and} \mybox{color0}{thoroughly} \mybox{color0}{delightful} \mybox{color0}{,} \mybox{color0}{it} \mybox{color0}{takes} \mybox{color0}{us} \mybox{color0}{on} \mybox{color0}{a} \mybox{color0}{roller} \mybox{color0}{-} \mybox{color0}{coaster} \mybox{color0}{ride} \mybox{color0}{from} \mybox{color0}{innocence} \mybox{color1}{to} \mybox{color0}{experience} \mybox{color0}{.}\\ % \mybox{color0}{[SEP]} \\
\fmergedstop &  \mybox{color0}{visually} \mybox{color0}{imaginative} \mybox{color0}{and} \mybox{color0}{thoroughly} \mybox{color0}{delightful} \mybox{color0}{,} \mybox{color0}{it} \mybox{color0}{takes} \mybox{color0}{us} \mybox{color0}{on} \mybox{color0}{a} \mybox{color0}{roller} \mybox{color0}{-} \mybox{color0}{coaster} \mybox{color0}{ride} \mybox{color0}{from} \mybox{color0}{innocence} \mybox{color1}{to} \mybox{color0}{experience} \mybox{color0}{.}\\
\midrule
\multicolumn{2}{l}{\bf NLI} \\
\multicolumn{2}{l}{\em\VanillaGrad} \\
\fop &  \mybox{color0}{a} \mybox{color0}{large} \mybox{color0}{,} \mybox{color0}{gray} \mybox{color1}{elephant} \mybox{color0}{walked} \mybox{color0}{beside} \mybox{color0}{a} \mybox{color0}{herd} \mybox{color0}{of} \mybox{color1}{zebra} \mybox{color0}{\#\#s} \mybox{color0}{.} \mybox{color0}{[SEP]} \mybox{color0}{the} \mybox{color1}{elephant} \mybox{color0}{was} \mybox{color1}{lost} \mybox{color0}{.} \mybox{color0}{[SEP]} \\
\fmergedstop &  \mybox{color3}{a} \mybox{color0}{large} \mybox{color0}{,} \mybox{color0}{gray} \mybox{color0}{elephant} \mybox{color0}{walked} \mybox{color0}{beside} \mybox{color2}{a} \mybox{color0}{herd} \mybox{color0}{of} \mybox{color0}{zebra} \mybox{color0}{\#\#s} \mybox{color0}{.} \mybox{color0}{[SEP]} \mybox{color1}{the} \mybox{color0}{elephant} \mybox{color0}{was} \mybox{color0}{lost} \mybox{color0}{.} \mybox{color0}{[SEP]} \\
\addlinespace 
\multicolumn{2}{l}{\em\SmoothGrad} \\
\fop &  \mybox{color0}{a} \mybox{color0}{large} \mybox{color0}{,} \mybox{color0}{gray} \mybox{color0}{elephant} \mybox{color0}{walked} \mybox{color0}{beside} \mybox{color0}{a} \mybox{color0}{herd} \mybox{color0}{of} \mybox{color0}{zebra} \mybox{color0}{\#\#s} \mybox{color0}{.} \mybox{color0}{[SEP]} \mybox{color0}{the} \mybox{color1}{elephant} \mybox{color1}{was} \mybox{color2}{lost} \mybox{color0}{.} \mybox{color0}{[SEP]} \\
\fmergedstop &  \mybox{color0}{a} \mybox{color0}{large} \mybox{color0}{,} \mybox{color0}{gray} \mybox{color0}{elephant} \mybox{color0}{walked} \mybox{color0}{beside} \mybox{color2}{a} \mybox{color0}{herd} \mybox{color0}{of} \mybox{color0}{zebra} \mybox{color0}{\#\#s} \mybox{color0}{.} \mybox{color0}{[SEP]} \mybox{color1}{the} \mybox{color0}{elephant} \mybox{color1}{was} \mybox{color0}{lost} \mybox{color0}{.} \mybox{color0}{[SEP]} \\
\addlinespace 
\multicolumn{2}{l}{\em\IntegratedGrad} \\
\fop &  \mybox{color0}{a} \mybox{color0}{large} \mybox{color0}{,} \mybox{color0}{gray} \mybox{color0}{elephant} \mybox{color0}{walked} \mybox{color0}{beside} \mybox{color0}{a} \mybox{color0}{herd} \mybox{color0}{of} \mybox{color0}{zebra} \mybox{color0}{\#\#s} \mybox{color0}{.} \mybox{color0}{[SEP]} \mybox{color0}{the} \mybox{color1}{elephant} \mybox{color0}{was} \mybox{color3}{lost} \mybox{color0}{.} \mybox{color0}{[SEP]} \\
\fmergedstop &  \mybox{color0}{a} \mybox{color0}{large} \mybox{color0}{,} \mybox{color0}{gray} \mybox{color0}{elephant} \mybox{color0}{walked} \mybox{color0}{beside} \mybox{color0}{a} \mybox{color0}{herd} \mybox{color0}{of} \mybox{color0}{zebra} \mybox{color0}{\#\#s} \mybox{color0}{.} \mybox{color3}{[SEP]} \mybox{color0}{the} \mybox{color0}{elephant} \mybox{color0}{was} \mybox{color2}{lost} \mybox{color0}{.} \mybox{color0}{[SEP]} \\
\midrule
\multicolumn{2}{l}{\bf Question Answering} \\
\multicolumn{2}{l}{\em\VanillaGrad} \\
\fop &  \mybox{color0}{Who} \mybox{color2}{caught} \mybox{color0}{the} \mybox{color4}{touchdown} \mybox{color1}{pass} \mybox{color0}{?} \mybox{color0}{[SEP]} \\
\fmergedstop &  \mybox{color0}{Who} \mybox{color0}{caught} \mybox{color7}{the} \mybox{color1}{touchdown} \mybox{color0}{pass} \mybox{color0}{?} \mybox{color0}{[SEP]} \\
\addlinespace 
\multicolumn{2}{l}{\em\SmoothGrad} \\
\fop &  \mybox{color0}{Who} \mybox{color3}{caught} \mybox{color0}{the} \mybox{color3}{touchdown} \mybox{color0}{pass} \mybox{color0}{?} \mybox{color0}{[SEP]} \\
\fmergedstop &  \mybox{color0}{Who} \mybox{color1}{caught} \mybox{color4}{the} \mybox{color0}{touchdown} \mybox{color0}{pass} \mybox{color0}{?} \mybox{color2}{[SEP]} \\
\addlinespace 
\multicolumn{2}{l}{\em\IntegratedGrad} \\
\fop &  \mybox{color6}{Who} \mybox{color0}{caught} \mybox{color0}{the} \mybox{color0}{touchdown} \mybox{color0}{pass} \mybox{color2}{?} \mybox{color1}{[SEP]} \\
\fmergedstop &  \mybox{color5}{Who} \mybox{color0}{caught} \mybox{color0}{the} \mybox{color0}{touchdown} \mybox{color0}{pass} \mybox{color2}{?} \mybox{color0}{[SEP]} \\
\midrule
\multicolumn{2}{l}{\bf Biosbias} \\
\multicolumn{2}{l}{\em\VanillaGrad} \\
\fop &  \mybox{color2}{she} \mybox{color0}{has} \mybox{color0}{had} \mybox{color0}{many} \mybox{color1}{years} \mybox{color0}{of} \mybox{color0}{experience} \mybox{color0}{and} \mybox{color0}{did} \mybox{color0}{thousands} \mybox{color0}{of} \mybox{color1}{operations} \mybox{color0}{.} \mybox{color1}{[SEP]} \\
\fmergedstop &  \mybox{color0}{she} \mybox{color0}{has} \mybox{color0}{had} \mybox{color0}{many} \mybox{color0}{years} \mybox{color0}{of} \mybox{color0}{experience} \mybox{color4}{and} \mybox{color0}{did} \mybox{color0}{thousands} \mybox{color2}{of} \mybox{color0}{operations} \mybox{color0}{.} \mybox{color0}{[SEP]} \\
\addlinespace 
\multicolumn{2}{l}{\em\SmoothGrad} \\
\fop &  \mybox{color1}{she} \mybox{color0}{has} \mybox{color0}{had} \mybox{color1}{many} \mybox{color1}{years} \mybox{color0}{of} \mybox{color1}{experience} \mybox{color0}{and} \mybox{color0}{did} \mybox{color1}{thousands} \mybox{color0}{of} \mybox{color0}{operations} \mybox{color0}{.} \mybox{color0}{[SEP]} \\
\fmergedstop &  \mybox{color0}{she} \mybox{color0}{has} \mybox{color0}{had} \mybox{color0}{many} \mybox{color0}{years} \mybox{color0}{of} \mybox{color0}{experience} \mybox{color1}{and} \mybox{color0}{did} \mybox{color1}{thousands} \mybox{color0}{of} \mybox{color1}{operations} \mybox{color0}{.} \mybox{color0}{[SEP]} \\
\addlinespace 
\multicolumn{2}{l}{\em\IntegratedGrad} \\
\fop &  \mybox{color4}{she} \mybox{color0}{has} \mybox{color0}{had} \mybox{color0}{many} \mybox{color0}{years} \mybox{color0}{of} \mybox{color0}{experience} \mybox{color0}{and} \mybox{color0}{did} \mybox{color0}{thousands} \mybox{color0}{of} \mybox{color0}{operations} \mybox{color0}{.} \mybox{color1}{[SEP]} \\
\fmergedstop &  \mybox{color5}{she} \mybox{color0}{has} \mybox{color0}{had} \mybox{color0}{many} \mybox{color0}{years} \mybox{color0}{of} \mybox{color0}{experience} \mybox{color0}{and} \mybox{color0}{did} \mybox{color0}{thousands} \mybox{color0}{of} \mybox{color0}{operations} \mybox{color0}{.} \mybox{color1}{[SEP]} \\
\bottomrule

\end{tabular}

\caption{Qualitative examples for all tasks and saliency methods when manipulating the gradient of \textit{stop words}. We show results before and after applying the \eviltwinU model. For QA, we only visualize the question. We omit \texttt{[CLS]} for space.}
\label{tab:qualitative_examples_stop_token}
\end{table*}

\end{document}